\pdfoutput=1

\documentclass[11pt]{article}

\usepackage[final]{acl}
\usepackage{times}
\usepackage{latexsym}
\usepackage{hyperref}
\usepackage{multirow}
\usepackage{booktabs}
\usepackage[T1]{fontenc}

\usepackage[utf8]{inputenc}

\usepackage{microtype}

\usepackage{inconsolata}

\usepackage{graphicx}

\title{Detection of Somali-written Fake News and Toxic Messages on the Social Media Using Transformer-based Language Models}

\author{\normalsize Muhidin A. Mohamed$^{1,*}$, Shuab D. Ahmed $^{2}$, Yahye A. Isse$^{2}$, Hanad M. Mohamed $^{2}$, \\ Fuad M. Hassan$^{3}$, Houssein A. Assowe$^{4}$ \\ \\
 $^1$Aston University, Birmingham, United Kingdom; $^2$Jamhuriya University, Mogadishu, Somalia;\\ $^3$Somali National University, Mogadishu, Somalia; $^4$ University of Djibouti, Balbala , Djibouti \\ \\
\texttt{Correspondence: m.mohamed10@aston.ac.uk}
}

\begin{document}
\maketitle
\begin{abstract}
The fact that everyone with a social media account can create and share content, and the increasing public reliance on social media platforms as a news and information source bring about significant challenges such as misinformation, fake news, harmful content, etc. Although human content moderation may be useful to an extent and used by these platforms to flag posted materials, the use of AI models provides a more sustainable, scalable, and effective way to mitigate these harmful contents. However, low-resourced languages such as the Somali language face limitations in AI automation, including scarce annotated training datasets and lack of language models tailored to their unique linguistic characteristics.
This paper presents part of our ongoing research work to bridge some of these gaps for the Somali language. In particular, we created two human-annotated social-media-sourced Somali datasets for two downstream applications, fake news \& toxicity classification, and developed a transformer-based monolingual Somali language model (named SomBERTa) -- the first of its kind to the best of our knowledge. SomBERTa is then fine-tuned and evaluated on toxic content, fake news and news topic classification datasets. Comparative evaluation analysis of the proposed model against related multilingual models (e.g., AfriBERTa, AfroXLMR, etc) demonstrated that SomBERTa consistently outperformed these comparators in both fake news and toxic content classification tasks while achieving the best average accuracy (87.99\%) across all tasks. This research contributes to  Somali NLP by offering a foundational language model and a replicable framework for other  low-resource languages, promoting digital and AI inclusivity and linguistic diversity.

\end{abstract}

\section{Introduction}

In today’s digital era, social media platforms have become a major source of news and an indispensable part of people’s lives for sharing opinions, debating on issues, connecting with networks, etc. Related to this, millions of Somali speakers use social networks including Facebook and Twitter to publicly express their thoughts and views about political and social events. However, with the ease of sharing unfactchecked and unedited information comes with unprecedented tide of biased news, misinformation, toxic content, among other issues. While fake news and toxic content problems are current global issues, the political instability and security problems in the horn of Africa region and the urge to know what is happening among the ordinary people, as well as the growing population of largely youthful Somali speaking social media users exacerbate the problems. This suggests the identification of different forms of Somali language written fake and toxic information on the social media utilizing state of the art AI technologies is not only paramount but will also contribute ot the security and well-being of the Somali communities in the Horn of Africa.   

On the other hand, the emergence of extensive pre-trained language models, such as BERT~\cite{devlin2018bert}, T5~\cite{raffel2020exploring}, and GPT-3~\cite{brown2020language}, have facilitated considerable advancements in high-resource languages such as English, and French. Despite these gains, low-resource languages (LRLs) including the Somali continue to be significantly underserved due to insufficient datasets, annotated corpora, and other vital linguistic resources. An estimated population of over 22 million people -- primarily in Somalia, Djibouti, Kenya, Ethiopia, and within Somali diaspora communities -- speak the langauge~\cite{mohamed2023lexicon, adelani2023masakhanews}. Although widely spoken, the Somali language -- just like other low-resource languages -- faces significant limitations in AI automation, annotated datasets, and the use of  NLP models tailored to its unique linguistic characteristics. This lack of resources hinders the development of effective AI tools, making tasks like machine translation, sentiment analysis, and text categorization challenging for Somali. Although multilingual models such as M-BERT~\cite{kenton2019bert,devlin2018bert}, AfriBERTa~\cite{ogueji2021small}, and AfroLM~\cite{dossou2022afrolm} aim to support low-resource languages, they often under-capture the Somali’s distinct syntax, semantics, and morphology due to their limited training data and vocabulary.


This research contributes to addressing the aforementioned challenges and gaps by compiling large diverse Somali pre-training corpus, creating human-annotated fine-tuning classification datasets, and developing a transformer-based Somali language model (SomBERTa) trained on these datasets and comparable in performance to multilingual models like AfriBERTa for Somali NLP tasks. In other words, the primary objective of this research is to create annotated fake news and toxicity Somali datasets from the social media and to develop a pre-trained Somali language model using diverse Somali text corpus then evaluate its performance against several other related state-of-the-art pre-trained models. The specific contributions of this paper are as follows: 
\begin{enumerate}
\item Compilation of a comprehensive and diverse Somali text corpus for training deep learning and large language models which can also serve as a valuable resource for future Somali NLP research.
\item Creation of labelled Social-media sourced Somali datasets for fake news and toxicity classification tasks 
\item Development of the first (to our knowledge) monolingual Somali large language model (SomBERTa), fine-tuned and evaluated on three different downstream tasks: fake news, toxicity, and news topic classification. 
\item Comparative performance analysis of the developed Somali specific LLM against several relevant state-of-the-art multilingual models including AfriBERTa and AfroXLMR.
\end{enumerate}

The creation of these resources form significant contributions for Somali NLP research which may also be adopted for other relevant low-resource languages. With the development of the pre-training and fine-tuning datasets, then training the Somali-specific base LLM and fine-tuning it followed with empirical evaluations, our research provides a replicable  methodological framework which can be easily adopted by other Somali (and/or even other low-resourced) AI/NLP researchers. The implementation code and data will be made available under academic license on Github\footnote{\url{https://github.com/shuakshay/FNaTCwSomBERTa}}.

\section{Related Literature}
\subsection{Pre-trained LLMs for low-resourced languages}
Pre-trained Large Language Models (PLLMs) leverage large-scale corpora to learn contextual representations that can be fine-tuned for various downstream tasks, revolutionizing NLP and enabling more efficient language processing~\cite{wang2023pre}.  PLLMs have lead to significance performance advances in various NLP tasks especially for high-resourced languages (e.g., English) but their application for low-resource languages including the Somali language is not well-studied~\cite{ojo2024good,adelani2023masakhanews}. One of the key challenges associated with PLLMs adoption for the Somali and other under-resourced languages is the scarcity of linguistic resources, such as extensive language specific corpora and labelled data for downstream tasks. Related to this, \cite{hedderich2020survey} provides a comprehensive survey of recent approaches addressing these challenges, highlighting techniques such as transfer learning and data augmentation. Similarly, \cite{siminyu2021ai4d, nekoto2020participatory} discuss relevant initiatives that focus on building resources and tools for African languages, emphasizing participatory research, community involvement and collaboration.

\subsection{Multilingual LLMs covering the Somali language}
Although the Somali NLP research has made some initial progress over the past few years in several tasks including text preprocessing~\cite{mohamed2023lexicon}, PoS tagging~\cite{mohammed2020using}, text classification~\cite{adelani2023masakhanews,alabi2022adapting}, machine translation~\cite{wang-etal-2024-afrimte, adelani-etal-2022-thousand}, and text retrieval~\cite{badel2023somali,adeyemi2024ciral}, yet the use of PLLMs for its various applications remain in its infancy stage compared to other languages. Generally, existing related research employing PLLMs can be roughly put into two categories: studies that focus on the development and pre-training of language models, and research works that emphasize fine-tuning models and downstream tasks including the creation of labelled datsets. 

The first research category explores the adoption of established LLMs for LRLs including the Somali language. For example, the work of~\cite{ogueji2021small} developed a multi-lingual LLM, called \emph{AfriBERTa}, for 11 low-resourced African languages including the Somali language. The training data for these languages was mainly sourced from the BBC News, and the resulting LLM was tested on two downstream tasks: named entity recognition and news topic classification. In another related landmark study~\cite{conneau2019unsupervised}, Facebook researchers introduced a comprehensive multilingual masked LLM (\emph{XLM-R}) which was trained on over 2-terabytes of Common Crawl data~\cite{wenzek2019ccnet}. The Somali was among 100 languages included in that LLM but with very small training data size of about 0.4 GP (62 million tokens). The model was evaluated on several NLP tasks including cross-linqual question answering and named entity recognition. Also, several related studies have adopted the aforementioned models (XLM-R, AfriBERTa) including the work of~\cite{alabi2022adapting} which applied multilingual adaptive fine-tuning for the Somali and 16 other African languages producing a tailored LLMs called \emph{AfroXLMR} (base, small, large). The AfroXLMR models were evaluated on three downstream tasks: NER, news topic classification and sentiment analysis. Moreover, \cite{wang2020extending} put forward \emph{E-MBERT}, an LLM that extended multilingual BERT (M-BERT)~\cite{kenton2019bert,devlin2018bert} to several low-resource languages not included in the original model such as the Somali language. The extension has been achieved by increasing the training vocabulary then re-training the LLM on the added languages.

The second class of the related works focused on labelled data creation and downstream tasks. This includes the work by~\cite{adelani2023masakhanews} which created news classification datasets and models for 16 African languages including the Somali. For the development of the classification LLMs, the study fine-tuned several existing multi-lingual text encoders including AfriBERTa-large~\cite{ogueji2021small}, XLM-R (Conneau et al., 2020), AfroXLMR~\cite{alabi2022adapting}, and AfroLM~\cite{dossou2022afrolm}. The Somali text included in fine-tuning the models comprised of 2,915 articles sourced from the BBC Somali (https://www.bbc.com/somali). Of these, AfriBERTa-large achieved the best Somali news classification accuracy of of 86.9 (F1-score). Besides, ~\cite{wang-etal-2024-afrimte} proposed machine translation evaluation metrics (AfriMTE and AfriCOMET) for 13 African languages which includes the Somali. In addition to several other contributions such as the creation of human evaluation dataset, the study used transfer learning techniques employing African-centric multilingual pre-trained language model such as AfroXLM-R-L~\cite{alabi2022adapting}. The somali translation pairs used in this research was extracted from the FLORES-200 dataset~\cite{nllb2022no}. Related to this, \cite{adelani-etal-2022-thousand} created news corpus translations for 16 African languages covering the Somali and adopted existing pre-trained models including low-resource tailored ones such as AfriBERTa to develop translation models. From their empirical investigation, the researchers found that fine-tuning pretrained LLMs using limited high quality translations can serve as an effective transfer learning method to new languages and domains. 
\subsection{Research gaps and our proposed work}
Advances in multilingual models, designed to support multiple languages within a single architecture (e.g., XLM-R, M-BERT, AfriBERTa, AfroXLMR, etc), achieved success in addressing NLP challenges for LRLs. While some of these models cover some African languages, including Somali, they often face limitations in adequately capturing language-specific syntax, grammar, and semantics due to minimal data representation during training~\cite{ogueji2021small,dossou2022afrolm}. For instance, a relatively limited Somali data was used in training most aforementioned multi-lingual LLMs which hinders their ability to accurately handle Somali-specific language tasks and produce good performance in downstream tasks. Also, there is an urgent need for annotated Somali corpora that can support the training of downstream applications, e..g, text classifications, machine translation, question answering, etc. In addition, monolingual models have shown greater success in capturing specific linguistic patterns when sufficient resources are available. Linked with this, research indicates that language-specific models, such as Arabic-BERT for Arabic~\cite{antoun2020arabert} and BETO for Spanish~\cite{canete2023spanish}, often outperform multilingual models on tasks requiring nuanced linguistic knowledge. 

Our study contributes to addressing these problems by compiling massive general purpose Somali corpus of about \emph{160 million} tokens, creating labelled datasets for fake news and toxicity classification tasks, and creating the first (to our knowledge) Somali-specific monolingual language model built on BERT~\cite{kenton2019bert} which we called \emph{ SomBERTa}. We believe that these constitute important contributions and valuable resources for further Somali NLP research.  
\section{Proposed transformer-based Somali language model}
In this section, we describe the research methodology for developing the proposed Somali-specific language model including the compilation of pre-trainig corpus, the creation of the fake news and toxicity datasets, and model training and evaluation.
\subsection{Model Develepment Datasets}
This section details the datasets utilized for pre-training and fine-tuning the proposed language model, SomBERTa. 
\subsubsection{Pre-training corpus}\label{PretraingCor}
To build a language-specific foundation model, we have compiled a large diverse Somali corpus covering news articles, literary works, social media content, and web-crawled text. These datasets were carefully curated from multiple sources to ensure that the proposed LLM is exposed to diverse and high-quality Somali texts. 
The key sources of our aggreated corpus were the Masakhane News Dataset (the Somali subset)~\cite{adelani2023masakhanews}, \href{https://www.bbc.com/somali}{BBC News Somali}, \href{https://t.me/somalibooks} {Somali Books}, \href{https://t.me/somalilibrary}{Somali Library}, AfriBERTa corpus (Somali subset)~\cite{ogueji2021small}, \href{https://metatext.io/datasets/cc100-somali}{CC100-Somali Dataset}~\cite{conneau2019unsupervised}, and CIRAL dataset (Somali subset)~\cite{adeyemi2024ciral}. These resources provided rich and comprehensive language coverage, essential for developing a robust language model and ensuring that it can be fine-tuned for various NLP contexts and downstream applications. \autoref{PreTrainDesc} provides a brief statistical description for each of the used pre-training datasets including the number of items (articles, books, etc.) and the size in sentences for each dataset. The combined pre-training corpus size made up a total of approximately \emph{160 million} tokens.

\begin{table}
  \begin{center}
  \caption{Statistical description of the pre-training Somali corpora including the number of items (articles, books, etc) and their sizes in sentences and tokens}\label{PreTrainDesc}
  \scalebox{0.80}
  {
  \begin{tabular}{llll}
    \hline
    \textbf{Dataset} & \textbf{\# of items} & \textbf{\# of sentences} & \textbf{\# of tokens} \\
    \hline
    Somali Books     & 75   & 136984 & 1776270 \\
    CIRAL-Somali     & 502852 & 2549809 & 64000292 \\
    CC100-Somali     & N/A  & 2399311 & 62331141\\
    BBC Somali     & 1260  & 32826 & 824122\\
    MasakaNews     & 1463  & 35902 & 836166\\
    SomAfriBERTa     & N/A  & 955114 & 30119511\\
    \hline
  \end{tabular}}
\end{center}
\end{table}

\subsubsection{Fake news and toxicity datasets} \label{FineTuningDatasets}
\begin{figure}[h!]
    \centering
   \includegraphics[width=0.95\linewidth]{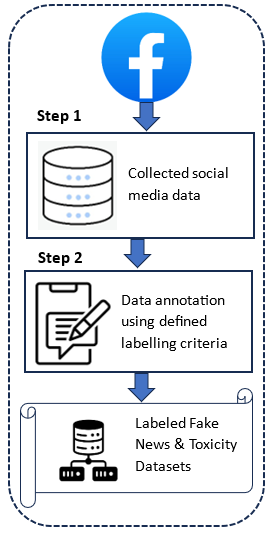}
    \caption{\textbf{Fake news and toxicity data collection and annotation}. For each dataset, we used a set of pre-defined annotation criteria that all annotators applied (see Appendix A)}
    \label{DataCollAnnot}
\end{figure}

In this study, we used Facebook as the social media platform from which the raw fake news and toxicity datasets were extracted. These two forms of data represented the core of the model fine-tuning and evaluation datasets among other classification tasks such as news categorization~\cite{adelani2023masakhanews}. Facebook was selected based on the fact that it is considered the most widely used social network for news consumption and online debating of political and social issues among the Somali speakers\footnote{https://gs.statcounter.com/social-media-stats/all/somalia}\footnote{ https://datareportal.com/reports/digital-2024-somalia}. In the context of this study, fake news and toxic messages are considered verifiable or identifiable Somali-written news and comments published on a Facebook channel by a news outlet, or other entities. \autoref{DataCollAnnot} illustrates the data collection, and annotation approach used in this work. The first step of the process involves extracting data from a diverse set of selected Facebook news channels. Then, in Step 2, the collected data are labeled by two annotators each.   

\begin{figure}[t]
  \includegraphics[width=\columnwidth]{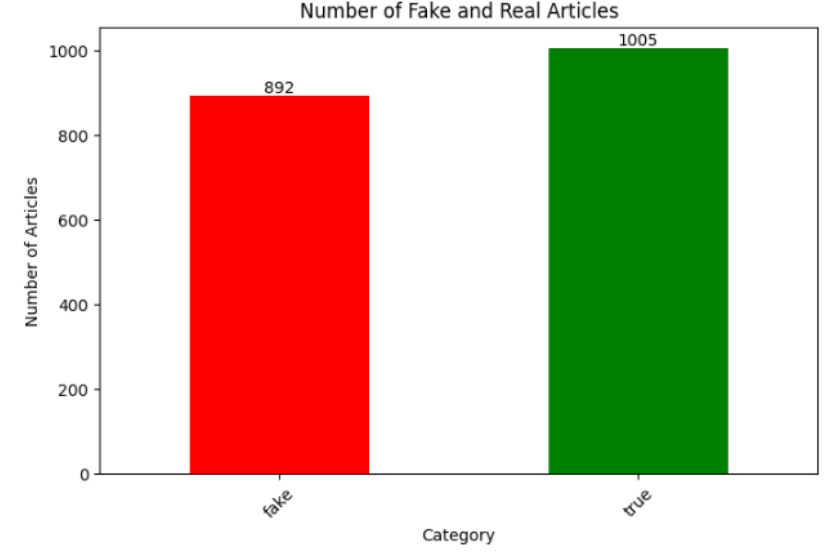}
  \caption{The distribution of the used fake and real news}
  \label{FakeNewsData}
\end{figure}

\paragraph{Fake news dataset:} \hspace{-0.4cm}
\autoref{tab1:fakenews}  summarizes the different Facebook source channels that were used for the collection of fake news data. They include \href{https://www.bbc.com/somali}{BBC News Somali} and \href{https://www.jowhar.com/}{Jowar}, two mainsources of authentic news and five other channels widely known to share false information among Somali speaking communities (these are anonymized for ethical purposes). Initially, a total of 3710 social media comments were collected for consideration including many comments that exhibited characteristics commonly associated with misinformation, such as sensational headlines, unverified sources, or misleading content. However, after a thorough labeling process involving fact-checking, cross-referencing with credible sources, and applying established fake news detection criteria (Appendix A), we filtered out stories that did not meet the threshold for being classified as fake/real. This refinement process reduced the total number of variable real/fake news instances to 1897 highlighting the importance of rigorous verification in distinguishing between real, and misleading fake news. \autoref{FakeNewsData} demonstrates the distribution of the annotated fake and real news as used in our study.  

\begin{figure}[h!]
    \centering
   \includegraphics[width=0.85\linewidth]{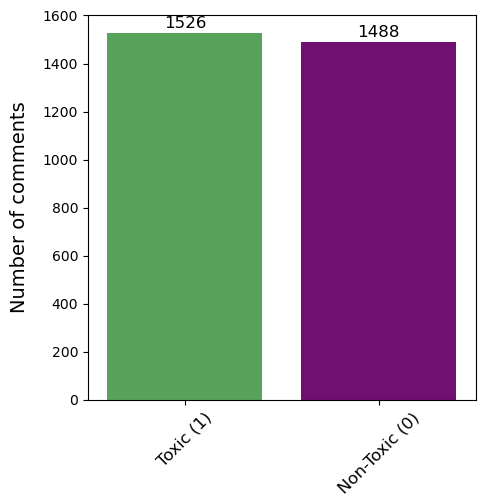}
    \caption{Toxic(1) and non-toxic(0) proportions of the toxicity dataset}
    \label{ToxicitData_binary}
\end{figure}

\begin{table}
  \centering
  \begin{tabular}{lc}
    \hline
    \textbf{Source channel} & \textbf{\# of posts} \\
    \hline
    \verb|BBC News Somali|     & {505}           \\
    \verb|Jowhar news site|     & {500}           \\
    \verb|Other channels|     & {892}           \\\hline
  \end{tabular}
  \caption{Souce Facebook channels of the fake news data and the number of posts retrieved from each}
  \label{tab1:fakenews}
\end{table}
\paragraph{Toxicity dataset:} \hspace{-0.4cm}
For the toxicity data, we have selected several controversial topics around political, social and legal issues, that were widely spoken and publicly debated by the Somali communities in the Horn of Africa (recent amendments to the Somali constitution, celebrity legal/court cases, Somalia's maritime dispute with neighbouring countries, etc). For each key post/video, we have collected all associated comments by the public for inclusion in the raw toxicity data. \autoref{tab:toxicity} shows each key topic and the number of associated annotated toxicity comments. We have applied a two-stage annotation process to categorize the comments. The first stage focused on a binary classification, where comments were labeled as either "toxic" or "non-toxic."  This stage was essential for distinguishing between harmful and neutral content. Two independent annotators reviewed each comment to ensure consistency and reliability in the labeling process. The second stage involved a further classification of the toxic comments into one or more of seven categories (multi-class): abuse, obscene, threat, insult, identity hate, severe toxic, and toxic. The annotation process employed in this study follows widely accepted toxicity classification principles applied in previous related research~\cite{saeed2021roman}.  \autoref{ToxicitData_binary} and \autoref{ToxicitData_multi} show the distributions of the binary and multiclass data partitions. It is clear that most of the toxic comments identified are predominantly of the identity-hate, insult and abuse categories (\autoref{ToxicitData_multi}). 


\begin{table}
  \centering
  \begin{tabular}{lc}
    \hline
    \textbf{Topic} & \textbf{\# of messages} \\
    \hline
    \verb|Constitutional Amendments|     & {1785}   \\
    \verb|Celebrity court case|     & {915}   \\
    \verb|Somali maritime case|     & {314}  \\\hline
  \end{tabular}
  \caption{Topics of the toxicity dataset and the number of posts from each topic}
  \label{tab:toxicity}
\end{table}

\begin{table}
  \centering
  \begin{tabular}{lc}
    \hline
    \textbf{Toxicity category} & \textbf{\# of messages} \\
    \hline
    \verb|Abuse|     & {585}           \\
    \verb|Obscene|     & {74}           \\
    \verb|Insult|     & {921}           \\
    \verb|Identity-hate|     & {868}           \\
   \verb|Severe-toxic|     & {203}           \\
    \verb|Threat|     & {216}           \\\hline
  \end{tabular}
  \caption{Identified toxicity data categoreis and the number of posts for each category}
  \label{ToxicitData_multi}
\end{table}


\subsubsection{Model development}
\begin{figure}[h!]
    \centering
   \includegraphics[width=0.98\linewidth]{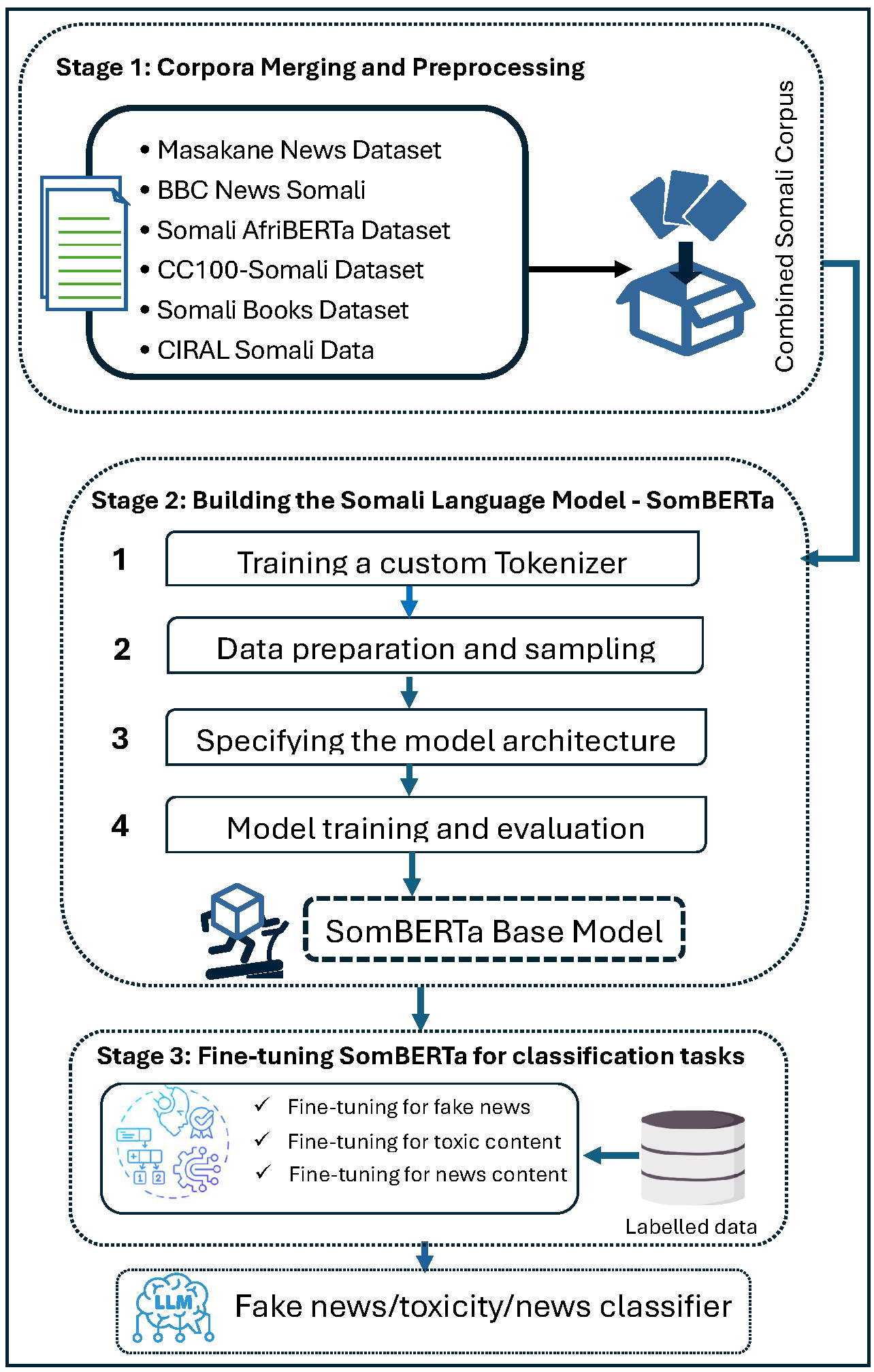}
    \caption{Architecture and implementation methodology of the proposed transformer-based Somali fake news and toxicity classification model}
    \label{SOMBERTa}
\end{figure}

Using the pre-traning and fine-tuning datasets described in sections~\ref{PretraingCor} and ~\ref{FineTuningDatasets}, the next step of our proposed language model was to build foundational Somali language model then refine for three downstream classification tasks, namely, fake news, toxicity, and news topic classifications. 
\autoref{SOMBERTa} illustrates the development pipeline of the proposed Somali language model which we named SomBERTa following the naming convention of LLMs built on BERT (Bidirectional Encoder Representations from Transformers)~\cite{devlin2018bert, kenton2019bert}. One of the reasons for selecting BERT model in our study is that it employs masked word prediction as a training strategy making it more suitable for text classification tasks (e.g., fake news detection, toxicity detection, etc) compared to other transformer-based models such as GPT~\cite{raschka2024build}. In fact, BERT is used by X (Twitter formerly) for the identification of toxic content. As shown in \autoref{SOMBERTa}, the implementation methodology of the SomBERTa model consists of three primary stages as described below. 
\begin{table}
  \centering
  \begin{tabular}{ll}
    \hline
    \textbf{Element} & \textbf{Description} \\
    \hline
     \verb|Title|     & {title of the writing, e.g. article, etc.}           \\
    \verb|Text|     & {content of the writing,  e.g. book, etc}           \\
    \verb|URL| & {address from which the writing is taken}  \\\hline
  \end{tabular}
  \caption{The three data elements used for the merged corpus}
  \label{DataElements}
\end{table}

\begin{figure}[t]
  \includegraphics[width=\columnwidth]{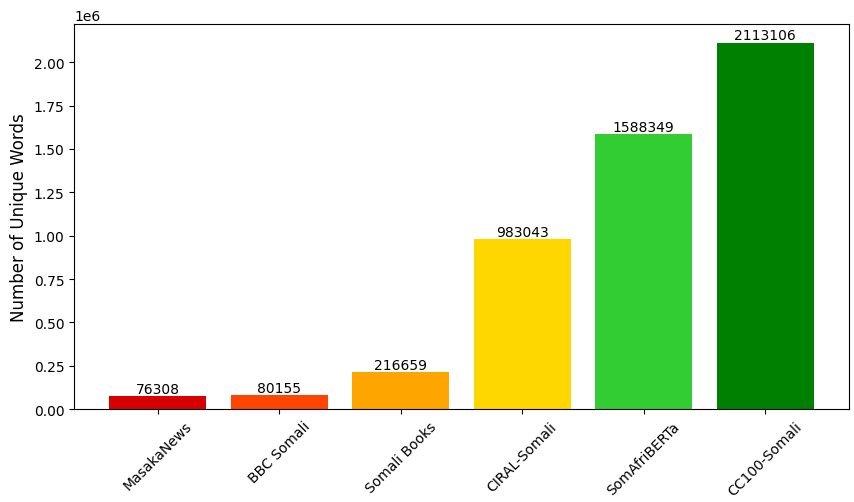}
  \caption{The unique words (vocabulary) for each of the merged datasets after preprocessing}
  \label{TrainDataDist}
\end{figure}

\begin{enumerate}
  \item \textbf{Corpora merging and preprocessing:} In the first stage, the six different datasets described in Section \ref{PretraingCor} have been merged and transformed to a shared template consisting of title-text-url trios (\autoref{DataElements}) for subsequent processing. 
  Next, a set of text cleansing and standardization steps (redundancy removal, character and whitespace filtering, lower-casing, etc) have been applied to prepare the data for model training and enable it capture non-noisy language patterns. 
  \autoref{TrainDataDist} shows the unique vocabulary sizes across the various datasets after preprocessing. Clearly, the size of the dataset vocabulary varies significantly with CC100-Somali and AfriBERTa datasets possessing the largest unique word counts followed by the CIRAL corpus.  The combination of these datasets provides diverse and expanded vocabulary coverage enabling our foundation LLM to learn syntax, semantics, and context of the Somali language.   
  \item \textbf{Building the Somali langague model:} Stage 2 is dedicated for the construction and pre-training of the proposed foundational model (SomBERTa) based on the corpus compiled and prepared in Stage 1. Key tasks of this stage included tokenizer training, data sampling for masked language modeling (MLM), and the specification of the model architecture, the training and evaluating the model.	SomBERTa's tokenizer was trained on WordPiece algorithm~\cite{song2020fast} on the rationale that the algorithm, initially popularized in BERT for its ability to handle complex linguistic structures through subword segmentation, is well suited for languages such as the Somali, where morphology includes intricate word formations with prefixes, suffixes, and compound structures that influence meaning~\cite{kenton2019bert}. In the tokenizer training, a vocabulary size of 70,000 tokens was selected to balance the model's learning of frequent terms and retention of rare subword components. 
  Next, data sampling and masking have been applied followed by dataset formatting through padding and truncation to standardize input sequences for efficient processing during training. The masking was used to enable SomBERTa gain a bidirectional context of the Somali text. Finally, BERT-based architecture -- customized to accommodate the specific vocabulary and structural requirements of the Somali language -- has been used to train the base model after which an intermediate evaluation of its prediction errors is performed using the MasakhaNews dataset~\cite{adelani2023masakhanews}. 

  \item \textbf{Fine-tuning SomBERTa for classification tasks:} The final stage of our research methodology and implementation constitutes fine-tuning the pre-trained base model (created in stage 2) for the three studied downstream tasks: toxic content detection, news classification, and fake news detection. This fine-tuning process employs labeled datasets for these classification datasets described in Section~\ref{FineTuningDatasets}. The hyperparameters for each task were carefully selected based on the recommendations in ~\cite{akiba2019optuna} to ensure that each configuration is optimized for prediction performance and efficiency.
\end{enumerate}
\begin{table*}[h!]
\begin{center}
\caption{Performance comparison of the proposed SomBERTa model with baseline and related multi-lingual models - all figures are percentage accuracy scores; $b$ and $m$ notations in the column headers respectively represent binary and multi-class classification }\label{tab:Comparators}%
\scalebox{0.92}{
\begin{tabular}{lllllll}
\hline
\toprule
\textbf{Model} & \textbf{Model Size} & \textbf{MasakhaNews(m)} & \textbf{Toxicity(b)} & \textbf{Toxicity(m)} & \textbf{FakeNews(b)} & \textbf{Average} \\
\midrule
\hline
SomBERTa$_{large}$	& 126 M	& 94.17 & 	\textbf{84.0} & \textbf{78.80} & \textbf{95.0} & \textbf{87.99} \\
BERT$_{base}$	& 110 M &  92.57 & 79.0 &  75.19 & 90.53 & 84.32\\
	AfriBERTa$_{large}$	& 126 M	& \textbf{95.24} & 	83.33 & 77.05 & 94.47 & 87.52 \\
	AfroXLMR$_{base}$ & 278 M & 94.77 & 81.67 & 75.63 & 92.63 & 86.18 \\
    RoBERTa$_{base}$ & 125 M & 85.71 & 77.5 & 68.47 & 92.11 & 80.95 \\
    DistillBERT$_{base}$ & 67 M & 92.66 & 80.33 & 76.78 & 90.53 & 85.08 \\
    ALBERT$_{base}$ & 12 M & 86.30 & 77.0 & 70.33 & 88.16 & 80.45 \\
    mBERT$_{base}$ & 168 M & 93.01 & 79.5 & 76.67 & 91.84 & 85.26 \\
\hline
\end{tabular}
}
\end{center}
\end{table*}
\section{Experimental evaluation}
To assess the performance of the proposed classification model, we bench-marked SomBERTa's fine-tuning performance against several other prominent related pretrained models, including BERT~\cite{devlin2018bert}, AfriBERTa~\cite{ogueji2021small}, AfroXLMR~\cite{alabi2022adapting}, DistilBERT~\cite{sanh2019distilbert}, RoBERTa~\cite{loureiro2022timelms}, ALBERT~\cite{lan2019albert}, and mBERT~\cite{kenton2019bert}. Each model is evaluated on the same task/dataset as SomBERTa:
\begin{itemize}
    \item \textbf{Binary Classification:} Fake news detection and binary toxic comment classification.
\item \textbf{Multi-Class Classification:} Somali news topic classification.
\item \textbf{Multi-Label Classification:} Multi-label toxic comment classification.
\end{itemize}
This benchmarking aims to evaluate SomBERTa's strengths and limitations relative to general-purpose multilingual models and other BERT variants on the selected Somali NLP tasks. For the performance quantification of the proposed classifier against compared models, we used the common classification metrics of accuracy, precision, recall, and F1-score. \autoref{tab:Comparators} presents a comprehensive comparison of SomBERTa’s performance against other models in terms of accuracy across the studied downstream tasks. Each model's average accuracy across all tasks is also included in the last column to provide an overall performance metric. From the results, SomBERTa demonstrates strong performance outperforming competing models in both fake news and toxic content classification tasks while achieving an average accuracy of 87.99\% across all tasks. Notably, SomBERTa ranks the third on the news classification task (MasakhaNews dataset) and is outperformed by both AfriBERTa and AfroXLMR models. This comparative analysis highlights that SomBERTa  is well-suited for Somali language tasks, performing competitively across diverse NLP applications, especially binary classification tasks. While some multilingual models like AfriBERTa, AfroXLMR, and mBERT also performed well, SomBERTa's Somali-specific pre-training allows it to capture unique language patterns that may contribute to its superior performance in certain tasks.

\section{Conclusion}
This paper presents a foundational step towards advancing NLP resources and models for the Somali language, a low-resource language with distinct linguistic characteristics. Through the collection of diverse Somali text corpus, the creation of labeled fake news and toxic content classification datasets, and the development of  SomBERTa -- a Somali-specific BERT-based model -- we have demonstrated that monolingual language models trained on targeted, diverse datasets can outperform general-purpose and multilingual models on several Somali NLP tasks. Obtained empirical evaluation results also highlighted the effectiveness and competitive performance of proposed transformer-based model in the identification of fake news, and toxic content. This underscores the potential of monolingual models for low-resource languages, especially when supported with a comprehensive and linguistically diverse corpus. This work not only contributes to addressing a critical gap in enabling Somali NLP but also offers a replicable framework for developing models for it and other underrepresented languages, promoting digital inclusivity and linguistic diversity of AI resources.

\section{Ethical considerations}
All datasets, including the pre-training and fine-tuning, used in this study are obtained from publicly available sources. The annotated fake news and toxicity datasets were gathered from public social media channels such as the pages of news and mainstream media outlets.  Any identifiable information in the associated commentaries has been redacted from the data with such excluded from the shared paper resources, e.g. implementation, data, etc, for the same purpose. Conclusions drawn from this work are based solely on identified patterns in the text. Our research adhered to ethical guidelines for public domain and social media data, and as such, we don’t see any potential harm or privacy risks from the use of the produced resources. 

\bibliography{acl_latex}
\section*{Appendices}
\appendix

\section{Data annotation}
We recruited a team of two volunteers who manually annotated the toxicity and fake news datasets, which were publicly gathered from social media. The annotators of the toxicity dataset followed a two-step annotation process to ensure accuracy and reliability in labeling.

For both tasks, the two annotators independently reviewed and labeled the toxicity comments as either "toxic" or "non-toxic" and the fake news text as either "fake" or "real" news. For the toxicity dataset, comments identified as toxic in the first stage were further categorized into seven predefined toxicity classes: "abuse," "obscene," "threat," "insult," "identity hate," "severe toxic," and "toxic." The categorization criteria for the toxicity dataset were based on widely accepted classification schemes used in prior research on toxic content detection, such as the work of~\cite{saeed2021roman}. To ensure the authenticity of real news content, we included news published by mainstream media outlets and trusted social media-based news pages. In addition, we incorporated various types of fake news instances to determine fake news labels following existing studies \cite{horne2017just,rashkin2017truth}. After annotation, each statement was assigned a final label only if both annotators agreed; otherwise, it was discarded.
\label{sec:appendix}

\end{document}